# Three multi-objective memtic algorithms for observation scheduling problem of active-imaging AEOS

Zhongxiang Chang [1,2※], Zhongbao Zhou[1,2]

[1]School of Business Administration, Hunan University, Changsha, 410082, China

[2]Hunan Key Laboratory of intelligent decision-making technology for emergency management, Changsha, China, 410082

## Abstract

Observation scheduling problem for agile earth observation satellites (OSPFAS) plays a critical role in management of agile earth observation satellites (AEOSs). Active imaging enriches the extension of OSPFAS, we call the novel problem as observation scheduling problem for AEOS with variable image duration (OSWVID). A cumulative image quality and a detailed energy consumption is proposed to build OSWVID as a bi-objective optimization model. Three multi-objective memetic algorithms, PD+NSGA-II, LA+NSGA-II and ALNS+NSGA-II, are then designed to solve OSWVID. Considering the heuristic knowledge summarized in our previous research, several operators are designed for improving these three algorithms respectively. Based on existing instances, we analyze the critical parameters optimization, operators evolution, and efficiency of these three algorithms according to extensive simulation experiments.

Keywords: Scheduling; Active imaging; OSWVID; Cumulative image quality; Multi-objective optimization; Memetic computing

## 1. Introduction

The imaging capability of earth observation satellites (EOSs) has been developed significantly since a new generation EOS named agile earth observation satellites (AEOSs) was designed and launched (Lemaître et al., 2002). AEOSs can modify their attitude angles (pitch, roll and yaw) during observation, which is named active imaging in our previous research (Chang et al., 2020). Active imaging is distinguished from the traditional image mode, named passive imaging, which needs holding the attitude angles during the whole observation (Wang et al., 2011). Since the capability of active imaging, as shown in Figure 1(a), AEOSs can observe a ground

---

※ Corresponding author: Zhongxiang Chang, E-mail address: zx_chang@163.com or zx_chang@hnu.edu.cn

target during its whole visible time window (VTW), therefore the image duration for observing the ground target is variable and belongs to $[p_s, d_i]$, where $p_s$ denotes the shortest image duration for every observation and $d_i$ represents the length of its VTW. While AEOSs without the capability of active imaging, as shown in the Figure 1(b), observe ground targets depended on AEOSs flight passively and the image duration of every ground target is given in advance and fixed in observation scheduling.

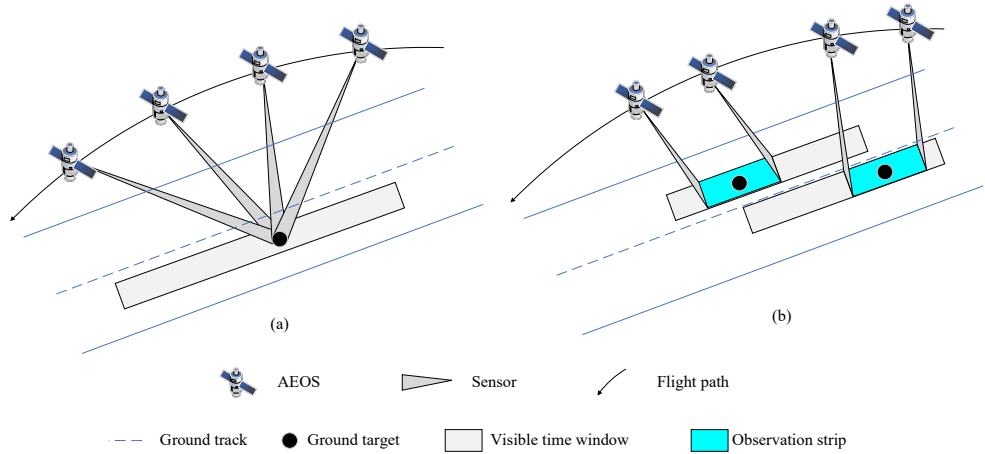

Figure 1 Active imaging and passive imaging

AEOS with the capability of active imaging has more observation choices for ground target, so the observation scheduling problem for AEOS with variable image duration (OSWVID) is more complex than the normal OSPFAS which is an NP-Hard combinatorial optimization problem (Lemaître et al., 2002) proved in theory .

OSWVID was analyzed in detail in our previous study (Chang et al., 2020, Chang et al., 2019). We designed extensive simulation experiments to analyze the relationship between the image duration of each ground target with its priority and workpiece congestion. By two types of tests, we summarized three heuristic knowledge.

**Knowledge 1** : When the workpiece congestion of task is constant, to achieve high overall profit, the high priority tasks should be allocated a longer imaging duration, while the low priority tasks should be assigned a shorter imaging duration.

**Knowledge 2** : The workpiece congestion of the fixed task much bigger is, more abandoned optional tasks will be with image duration of the fixed task increasing

**Knowledge 3** : On the premise that the image duration of every observation task is equal when the standard imaging duration ($\overline{tp}$) of each ground target belongs to $[0.5, 0.9]$, the scheduling benefit achieves the maximum value.

The efficiency of the three heuristic knowledge is discussed by the controlled trial considering extensive simulation experiments. Though adopting three pieces of knowledge into the greedy algorithm proposed in our previous research improved the algorithm to some extent, the previous research was too theoretical.

(1) The transition time was constant and can be calculated according to the location of the ground target in advance, which may be reasonable for problem analysis, but is significantly not suitable for further research. Frankly speaking, the transition time for AEOSs is time-dependent (Liu et al., 2017). We will propose the calculation equation for the transition time of AEOSs with the capability of active imaging in section 2.

(2) We did not consider the image quality of every ground target in the previous. For AEOSs without the capability of active imaging, the image quality of each ground target is depended on the observation start moment (He et al., 2019a, He et al., 2018, Peng et al., 2020, Peng et al., 2019, Liu et al., 2017), while for AEOSs with the capability of active imaging, the image quality of each ground target will become more complex and is a cumulative variable, and we will discuss it in section 3.2.

(3) The energy consumption was also not considered in our previous paper. We assumed the power is enough. To further research, energy consumption should be considered more practically. A definition of detailed energy consumption then will be also proposed in section 3.3.

(4) We only proposed a simple greedy algorithm to solve the problem and did not compare it with other existing algorithms.

In this study, we will consider cumulative image quality, detailed energy consumption, and time-dependent transition time to research OSWVID further. The several major contributions of this paper are summarized below.

(1) A novel function is proposed to calculate the image quality for AEOSs with the capability of active imaging.

(2) We consider two types of energy consumption during observation of AOESs and build an overall function to calculate the energy consumption.

(3) Considering the image quality and energy consumption, we model OSWVID as a bi-objective optimization problem.

(4) Toward to characteristics of OSWVID, we propose three multi-objective optimization algorithms, PD+NSGA-II, LA+NSGA-II, and ALNS+NSGA-II, based on NSGA-II.

(5) Based on the heuristic knowledge summarized in our previous research, several operators are designed for improving these three algorithms respectively.

The paper is structured as follows. Section 2 presents the review of related literature. In section 3, two optimization objectives are designed in detail and then a bi-objective optimization is built for describing OSWVID. Section 4 proposes three algorithms, PD+NSGA-II, LA+NSGA-II and ALNS+NSGA-II based on the nondominated sorting genetic algorithm II (NSGA-II), and several operators are designed according to the heuristic knowledge summarized in our previous research. Section 5 reports on the experimental results and offers a detailed performance analysis. Section 6 concludes this paper and identifies future research directions.

## 2. Literature reviews

The research (Wolfe and Sorensen, 2000) is very similar to our study. They saw the image duration of each ground target belonged to $[d_{min}, d_{max}]$, $d_{min}$ and $d_{max}$ denotes minimum image duration and maximum image duration respectively. They proposed a trapezoidal suitability method, by drawing not a mathematical function, to define the benefits of different observation windows (OWs) during the whole visible time window (VTW). And they designed three types of algorithms, priority dispatch algorithm (PD), look ahead algorithm (LA) and genetic algorithm (GA), to solve the problem. However, the problem in their paper was more simplified, they did not consider the transition time between different observation tasks and energy consumption, which may be right in 2000. But now, the ground targets are redundant (Wang et al., 2016) according to the imaging capability of AEOS, and the transition time between different observations is time-dependent (Liu et al., 2017) and on-board power is not enough for infinite observations (Zhao et al., 2019). These situations inspire us to have to study OSWVID further in our paper.

After Wolfe (Wolfe and Sorensen, 2000), there were a few types of research studied observation scheduling problems for AEOS with variable image duration (OSWVID) directly according to our limited literature searching.

Bunkheila (Bunkheila et al., 2016) analyzed the mission planning problem of AEOS in detail, simplified the problem into two sub-problems: geometry analysis to strip the ground targets and temporal analysis to calculate the visible time windows (named as data take opportunities in their research). It's obvious that AEOS studied in their paper can image actively. But actually, the problem in their research cannot be named OSPFAS, because they focused on the acquisition operations and did not consider any constraint between different ground targets or tasks.

Cui (Cui et al., 2018) studied the mission planning of video imaging satellites, established a constraint satisfaction model using short-period planning (1/6 orbit), and solved the problem using an ant search algorithm with the taboo list. However, they simplified the transition time as a constant variable without time-dependent, and in simulation experiments, they proposed two fixed image duration for observing every ground target without any theoretical analysis.

Nag (Nag et al., 2018) extended the image duration for each unique image in simulation experiments, but according to their model, they could not solve observation scheduling with the variable image duration for each unique image. And Nag (Nag et al., 2019) just transformed their previous study (Nag et al., 2018) on board, there was not any change in algorithm or modeling. Given all that, it is lacked necessary theoretical analysis in their researches to achieve Observation scheduling problem for AEOS with variable image duration.

It is a pity that AEOS(s) proposed in most existing researches is(are) not provided with the capability of active imaging. The image duration for every ground target/task is fixed and given in advance. Considering the area of ground targets, the recent researches can be classified into two kinds: small ground targets (point tasks or small areas) that AEOS can be observed in one pass (Wang et al., 2015, Liu and Yang, 2019, Xu et al., 2016, Augenstein et al., 2016, Wu et al., 2019, He et al., 2019b, Wang et al., 2019b, Shao et al., 2018, Gabrel et al., 1997, Chang et al., 2022), and large ground targets (Niu et al., 2018, Berger et al., 2020, Xu et al., 2020). In this paper, we only consider small ground targets.

Some authors ignored the transition time (Wolfe and Sorensen, 2000, Tangpattanakul et al., 2015) of AEOSs or saw it as a constant variable (Hu et al., 2019, Lemaître et al., 2002, Wang et al., 2019a, Chang et al., 2020). Actually, more and more authors saw the transition time of AEOSs as a time-dependent variable (He et al., 2019a, Wang et al., 2019c, Li and Li, 2019, He et al., 2018, Peng et al., 2020, Peng et al., 2019, Liu et al., 2017, Lu et al., 2021, Chang et al., 2021b, Chang et al., 2021a), because of the flexible attitude maneuver of AEOSs. They proposed several definitions for calculating the transition time, but actually, all of them is the same. Like the calculation equation proposed in the study (Liu et al., 2017) shown as following.

$$trans(\Delta g) = \begin{cases} 35/3 & \Delta g \leq 10 \\ 5 + \frac{\Delta g}{v_1} & 10 < \Delta g \leq 30 \\ 10 + \frac{\Delta g}{v_2} & 30 < \Delta g \leq 60 \\ 16 + \frac{\Delta g}{v_3} & 60 < \Delta g \leq 90 \\ 22 + \frac{\Delta g}{v_4} & \Delta g > 90 \end{cases} \quad (1)$$

where the unit of transition time is second. $v_1$, $v_2$, $v_3$ and $v_4$ are four different angular transition velocities. The values of them are $v_1 = 1.5°/s$, $v_2 = 2°/s$, $v_3 = 2.5°/s$ and $v_4 = 3°/s$. $\Delta g$ denotes the total change of attitude angles between two observations and is calculated by the function (2).

$$\Delta g = \Delta \pi + \Delta \gamma + \Delta \psi \quad (2)$$

where $\Delta \pi$, $\Delta \gamma$, and $\Delta \psi$ indicate respectively the change of the pitch angle, the roll angle, and the yaw angle. The attitude capability of AEOS studied in our paper is much greater than that of AOES in the existing researches, but the attitude maneuver is completely the same ignoring the angular transition velocities. On the other hand, there is no in-orbit satellite with the capability of active imaging in China and we do not know the exact parameter of the angular transition velocities, so we adopt the calculation equation, the function (1), directly in our research.

## 3. Mathematical formulation

In this section, some assumptions will be proposed to simplify OSWVID reasonably. Two equations then are designed for calculating the image quality and energy consumption. Given all that, a bi-objective optimization model is built considering several constraints to describe OSWVID.

### 3.1. Preliminaries

To better solve the focus in our paper, several reasonable assumptions are proposed to standardize and simplify OSWVID referring to previous researches and engineering experiences.

(1) The yaw angle reflects the image quality significantly, but correcting the yaw angle is a focus for attitude scheduling (Wu et al., 2020) not for observation scheduling. So, the yaw angle will be sawed as a constant variable in our research.
(2) We do not pay attention to the attitude maneuver during observation, therefore, we assume that the attitude maneuver is feasible during the whole VTW.
(3) Since different types sensors have different attributions, only a single optical AEOS is considered in our study.
(4) AEOS has sufficient on-board memory during the whole scheduling horizon, it is not necessary to consider satellite image data downlink.
(5) We only consider small ground targets that AEOS can be observed in one pass. Because large ground targets need to be decomposed before task scheduling (Niu et al., 2018, Berger et al., 2020, Xu et al., 2020), which is irrelated to the focus in our paper.
(6) Since ground targets are redundant (Wang et al., 2016) according to the image capability of AEOS, we assume every ground target can be observed at most once.
(7) We assume the sensor is on during the whole scheduling horizon. By the way, it is not necessary to spend extra time and energy for opening the sensor before every observation.

Based on the aforementioned assumptions, the observation scheme $S$ for OSWVID can be generally described by:

$$S = \{Sat, St, Et, AGT\} \quad (3)$$

where the included notations are defined as follows:
- $Sat = \{sat, |S| = n_s\}$ is the set of satellites for observation scheduling. In this paper, we just consider a single optical AEOS, so we can ignore this attribute.
- $[St, Et]$ is the scheduling horizon for OSWVID.
- $AGT = \{gt_i | 1 \leq i \leq n_{gt}, |AGT| = n_{gt}\}$ indicates the set of all ground targets considered in the observation scheduling, $n_{gt}$ means the quantity of all ground targets. Each ground

target $gt_i$ is composed of a six-tuple:

$$gt_i = \{Id, \omega, d_0, c_0, w, ow\} \tag{4}$$

where $Id$ is the identifier of the ground target $gt_i$, $\omega$ reflects the priority of $gt_i$, $d_0$ is the requirement image duration of $gt_i$. $c_0$ is the original workpiece congestion of $gt_i$ as defined in the function (26). $w$ denotes the visible time window (VTW) of $gt_i$ and is defined as function (5), and $ow$ reflects the observation window (OW) for observing $gt_i$ defined as the function (6) and $ow \subset w$.

$$w = \{Id, s, e, b_0, rList, pList\} \tag{5}$$

where $Id$ is the identifier of $w$ for distinguishing several VTWs[1] of a same ground target, $s$ and $e$ represents begin time and end time of $w$ respectively, $b_0$ is an observation moment with the highest instant image quality during $w$ and calculated by the function (9). $rList$ and $pList$ indicates the set of roll angles and pitch angles on different observation moment during $w$ respectively.

$$ow = \{b, e, d, p_o, r_o, p_\infty, r_\infty\} \tag{6}$$

where $b$ and $e$ denotes the true begin moment and the true end time of $ow$, $d$ reflects the true image duration of $ow$. $p_o$, $r_o$, $p_\infty$ and $r_\infty$ indicates the begin pitch angle, begin roll angle, end pitch angle, and end roll angle of $ow$ respectively.

### 3.2. Calculate image quality

Without the capability of active imaging, the image quality is only depended on the observation start moment (Liu et al., 2017, Peng et al., 2019, Peng et al., 2020, He et al., 2018, He et al., 2019a), because AEOSs will hold attitude angles during the whole OW. Without loss of generality, let the calculation function proposed in Peng (Peng et al., 2020) as an example. They referenced Liu (Liu et al., 2017) and He (He et al., 2018) defined a function to calculate the image quality shown as the function (7). The image quality ($q$) belonged to the interval [0,1] and depended on the observation pitch angle directly.

$$q(u) = 1 - \frac{|\pi(u)|}{90} \tag{7}$$

where $\pi(u)$ denotes the pitch angle when AEOS observes the ground target at the observation start moment $u$.

As mentioned above, the image quality in the function (7) is only dependent on the begin pitch angle, while AEOSs with the capability of active imaging can modify the attitude angles during the whole OW. Therefore, the image quality for AEOSs with the capability of active imaging should be redefined as a cumulative variable, defined as the function (8), and is related to the pitch angles and roll angles during the whole OW.

---

[1] A ground target with several VTWs will be clone as several ground targets with the same identifier but different VTWs.

$$Q(ow) = \frac{\sum_{u \in ow} q(u)}{\sum_{u \in VTW} q(u)} \quad ow \subset VTW \tag{8}$$

where $ow$ and $VTW$ denotes observation window and visible time window respectively. $q(u)$ represents the instant image quality obtained at the moment $u$, which is redefined as the function (9). $\sum_{u \in ow} q(u)$ reflects the true image quality obtained by observation, while $\sum_{u \in VTW} q(u)$ indicates the maximum image quality under observation in the whole VTW. By the way, $Q(ow)$ belongs to the interval [0,1] and the bigger value of it is, better is.

$$q(u) = \left(1 - \frac{|\pi(u)|}{90}\right) \times \left(1 - \frac{|\gamma(u)|}{90}\right) \tag{9}$$

According to the assumption (1), the yaw angle is constant in our research. $\pi(u)$ and $\gamma(u)$ denotes the instant pitch angle and instant roll angle respectively when AEOS observes the ground target at the moment $u$. In addition, the instant image quality ($q(u)$) belongs to the interval [0,1]. And the value of $q(u)$ bigger is, the instant image quality better is.

### 3.3. Calculate energy consumption

Since the assumption (7), two types of energy consumption, energy consumption by observation (EO) and energy consumption by attitude conversion (EC), should be considered in our study.

To facilitate to calculate these two types of energy consumption for an observation scheme $S$, some notations and symbols are defined as below.

- $E$      total energy consumption of $S$
- $ot$      total observation time of $S$
- $ct$      total attitude conversion time in $S$
- $eo$      energy consumption rate for observation
- $ec$      energy consumption rate for attitude conversion
- $GT$      the set of ground targets scheduled to be observed

The total energy consumption of the observation scheme $S$ is the sum of EO and EC and is defined as the function (10).

$$E = eo \times ot + ec \times ct \tag{10}$$

where $eo$ and $ec$ are constant variables, that is, $eo = 0.08W$ and $ec = 0.05W$. The calculation function of $ot$ and $ct$ are calculated respectively following.

$$ot = \sum_{i=1}^{|GT|} gt_i.ow.d \tag{11}$$

where $|GT|$ denotes the quantity of ground targets scheduled to be observed. $gt_i.ow.d$ is the true image duration for observing ground target $gt_i$.

$$\Delta g_{gt_i \to gt_{i+1}} = |gt_{i+1}.ow.p_o - gt_i.ow.p_\infty| + |gt_{i+1}.ow.r_o - gt_i.ow.r_\infty| \tag{12}$$

Let $\forall gt_i, gt_{i+1} \in GT$ as examples, which are two adjacent scheduled ground targets in $\mathcal{S}$[2]. $\Delta g$ in the function (1) can be calculated by the function (12). Regarding the assumption (1), the yaw angle is constant. So we just consider the change of pitch angle and roll angle.

So, $ct$ in $\mathcal{S}$ can be calculated by

$$ct = \sum_{i=1}^{|GT|-1} trans(\Delta g_{gt_i \to gt_{i+1}}) \tag{13}$$

## 3.4. A bi-objective optimization model

Three decision variables, $x_i$, $gt_i.ow.b$ and $gt_i.ow.d$, are considered in the model. $x_i$ represents whether ground target $gt_i$ is scheduled to be observed. While $gt_i.ow.b$ denotes the observation begin moment of $gt_i$, which is belonged to its VTW ( $[gt_i.w.s, gt_i.w.e]$ ). $gt_i.ow.d$ indicates the true observation duration of $gt_i$, the value range of which can be calculated by the method proposed in our previous study (Chang et al., 2020).

Observing as much ground targets as possible is the original intention of Observation scheduling problem for (A)EOS. Without loss of generality, we design the minimum loss rate of image quality (LR) as one optimization objective in our study, which is similar to maximize the sum of image quality of scheduled ground targets.

$f_1(\mathcal{S})$, the loss rate of image quality (LR), namely, LR considers the priority and image quality of scheduled ground targets and is belonged to the interval [0,1].

$$f_1(\mathcal{S}) = 1 - \frac{\sum_{i=1}^{n_t} x_i \times gt_i.\omega \times Q(gt_i.ow)}{\sum_{j=1}^{n_{gt}} gt_j.\omega} \tag{14}$$

where $\sum_{j=1}^{n_{gt}} gt_j.\omega$ is a constant variable, which is calculated under the assumption that all ground targets are observed during their whole VTW. $\sum_{i=1}^{n_t} x_i \times gt_i.\omega \times Q(gt_i.ow)$ denotes the sum of priority and image quality of scheduled ground targets in their OW.

Spending less energy is an apparent goal for observation scheduling (Wang et al., 2015). Minimizing the energy consumption (EC) is not only good for observation, but also is well for the working of AEOS system. So we consider EC as another optimization objective.

$$f_2(\mathcal{S}) = \frac{E}{MEC} \tag{15}$$

where $E$ denotes the total energy consumption to observe all scheduled ground targets as defined in the function (10). To normalize $E$, a constant variable, named maximum energy consumption (MEC), is proposed. MEC is calculated under the assumption that all ground targets are observed during their whole VTW. After the normalization, $f_2(\mathcal{S})$ also belongs to the interval [0,1].

---
[2] Notably, all scheduled ground targets are sorted by their true observation begin moment ascending.

$$MEC = \sum_{j=1}^{n_{gt}} \left(eo \times (gt_j.w.e - gt_j.w.s)\right) + maxTrans \times ec \times n_{gt} \quad (16)$$

where $maxTrans$ denotes the maximum transition time, and according to the function (1), we set $maxTrans = 100$. $(gt_j.w.e - gt_j.w.s)$ denotes the length of VTW.

$$min\ F(S) = \{f_1(S), f_2(S)\} \quad (17)$$

Given all that, we obtain two optimization objectives, the loss rate of image quality (LR) and the energy consumption (EC), as shown in the function (17). These two objectives are not irreconcilable in observation scheduling, so the dual optimization of them is reasonable and possible. OSWVID is a bi-objective discrete optimization problem (Kidd et al., 2020). Then we will propose all constraints of OSWVID.

$$x_i \leq 1 \quad 0 \leq i \leq n_{gt} \quad (18)$$

The constraint (18) indicates all ground targets are observed at most once, which is consistent with the mention in the assumption (6).

$$\begin{cases} gt_j.ow.b \geq gt_j.w.s \\ gt_k.ow.e \leq gt_j.w.e \end{cases} \quad 0 \leq j \leq n_{gt} \quad (19)$$

The constraint (19) means all ground targets must be observed during their VTW and the image duration ($gt_j.ow.d$) cannot exceed the length of VTW.

$$gt_i.ow.d \geq x_i \times gt_i.ow.d_0 \quad 0 \leq i \leq n_{gt} \quad (20)$$

If $x_i = 1$, the true image duration of $gt_i$ cannot be smaller than its requirement image duration. Otherwise, this constraint is invalid. Combining the function (19) and (20) represents the complete constraint for confirming OW of each ground target.

$$\begin{cases} x_j x_k \times (gt_k.ow.b - gt_j.ow.b) \leq x_j x_k & \forall k,j \in [0, n_{gt}] \\ x_j x_k \times (gt_j.ow.b - gt_k.ow.e) \geq x_j x_k \times trans\left(\Delta g_{gt_k \to gt_j}\right) \end{cases} \quad (21)$$

The two equations in the constrain (21) impose the transition time constraint. If only if both of $x_k$ and $x_j$ equal 1, the first function means $gt_k$ and $gt_j$ are two arbitrary observed ground targets, and $gt_k$ is observed before $gt_j$. The second function represents the interval time between them must be bigger than the attitude conversion time from $gt_k$ to $gt_j$. Otherwise, if any of $x_k$ or $x_j$ equals zero, the constrain (21) will be invalid.

## 4. Three memetic algorithms

Memetic computing/algorithm (MC/MA) usually (Neri and Cotta, 2012) consists of an evolutionary framework and a set of local search algorithms, which are activated within the generation cycle. This algorithmic framework has been successfully used in solving various combinatorial optimization problems(Galinier et al., 2011).

We design three adaptive bi-objective memetic algorithm, called ALNS+NSGA-II, PD+NSGA-II and LA+NSGA-II, to solve OSWVID. They combine the corresponding breeding algorithms with a nondominated sorting genetic algorithm II (NSGA-II). The basic framework of the algorithm is given as a flow chart in Figure 2.

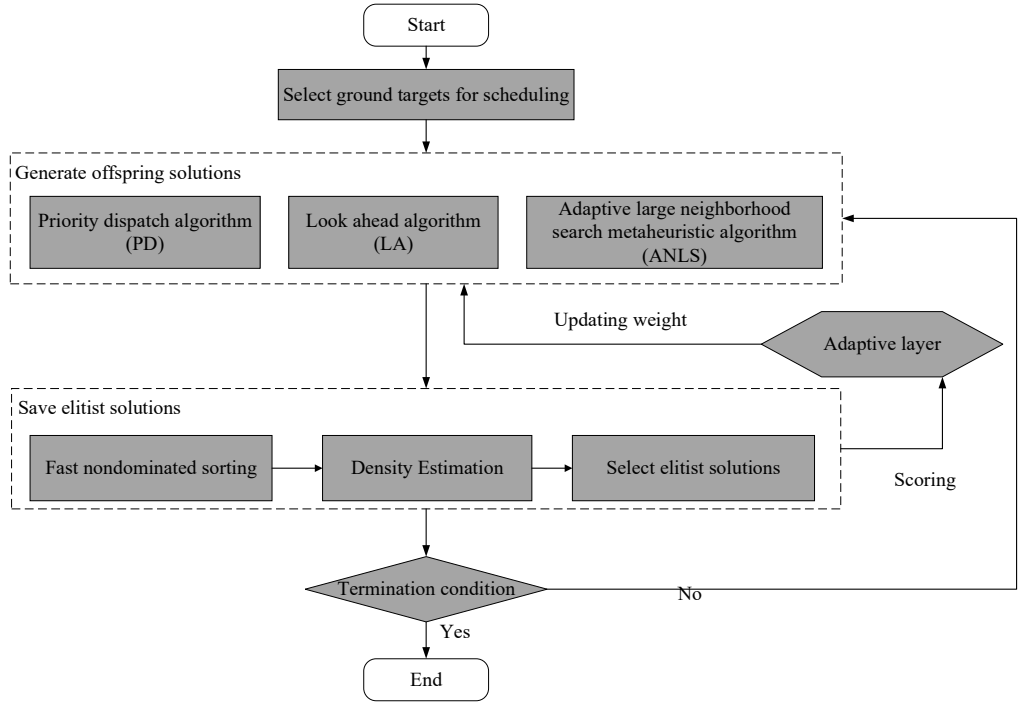

Figure 2 The framework of our algorithms

Three breeding algorithms, priority dispatch algorithm (PD), look ahead algorithm (LA) and adaptive large neighborhood search metaheuristic algorithm (ALNS), are designed for generating offspring solutions. PD and LA were proposed in the research (Wolfe and Sorensen, 2000), while ALNS was from Liu (Liu et al., 2017). These three algorithms have their advantages respectively. Note that, we will redesign them according to the characteristic of OSWVID in follows.

NSAG-II is adopted as the evolutionary mechanism for our three algorithms to achieve Pareto frontier. Deb et al. (Deb et al., 2002) improved NSGA (Srinivas and Deb, 1995) and proposed NSAG-II in 2002, which is one of the best known evolutionary multi-objective optimization algorithms (Gong et al., 2009). Furthermore, NSGA-II has been adopted for several evolutionary computing paradigms(Wu and Che, 2019, Kadziński et al., 2017, Wu and Che, 2020).

In addition, our algorithms will adopt the famous box-method (Hamacher et al., 2007), which is based on $\varepsilon$-constraint method, to guide the evolution in the non-dominated space.

## 4.1. Select ground targets

Since the ground targets are redundant (Wang et al., 2016) according to the image capability of AEOS, we do not consider whole ground targets during each scheduling. Ground targets are selected randomly. We let $RS \in [0,1)$ to control the selecting process. If only if the random possibility is bigger than $RS$, ground target will be selected. We will analyze effect of different value of $RS$, and suggest suitable value of $RS$ for different algorithms.

## 4.2. Priority dispatch algorithm (PD) and Look ahead algorithm (LA)

The priority dispatch algorithm (PD) and the look ahead algorithm (LA) were proposed in the research (Wolfe and Sorensen, 2000). Both of the PD algorithm and the LA algorithm are greedy constructive algorithms and do not consider backtracking. They are easy to understand because they use three simple phases: Sorting, Scheduling, and Expanding, while the three phases were named as: Selecting, Allocation, and Optimization in the research (Wolfe and Sorensen, 2000). The LA algorithm is similar to the PD algorithm but it uses a smarter Scheduling step. The solution obtained by Scheduling of LA at any time is a locally optimal solution toward considered ground targets. The basic structure of the PD algorithm and the LA algorithm is similar, so let PD as an example, the flow of PD is shown in Algorithm 1.

Algorithm 1 **:** The priority dispatch algorithm (PD)

**Input:** A set of selected ground targets $IGT$
**Output:** A solution $S$, the element of which is a set of ground targets $GT$ which are scheduled to be observed

1: **Sort** all selected ground targets in the set $IGT$
2:     Choose a **sort operator** according to the adaptive layer
3:     Sorting all selected ground targets by guidance information of selected **sort operator**
4: **Stop Sort**
5: **Schedule** all sorted ground targets
6:     Allocate its requirement image duration ($d_0$) of each target;
7:     Adopting **dichotomization** to decide OW for each ground target
8:     Justify whether the ground target can be observed and add the successful one into $GT$
9: **Stop** until all selected ground targets in $IGT$ are traversed
10: **Expand** the image duration of all scheduled ground targets
11:     Choose a **expand operator** according to the adaptive layer
12:     Adopting **dichotomization** to expand observation begin time and observation end time of ground target respectively one by one
13: **Stop** until all scheduled ground targets in $GT$ are traversed
14: **Output** the solution $S$

Note that, if line 6: is enriched considering variable image duration, Algorithm 2 will propose the LA algorithm. It traverses all image duration $d \in [d_0, w.e - w.s]$ for every ground target from big value to small value until scheduling success. So, the solution obtained by the scheduling of LA at any time is a locally optimal solution toward considered ground targets, but this also results in LA consumes more runtime. And there are three important elements, **sort operator**, **dichotomization scheduling,** and **expand operator**, in Algorithm 1. We will talk about them respectively in detail following.

### 4.2.1. Sort operators

In our implementation, four sort operators are defined according to different guidance information. Two of them, **P-Sort** and **C-Sort**, are designed according to our previous research (Chang et al., 2020).

**R-Sort**: The guidance information of this operator is generated randomly. Then all selected ground targets are ranked by their guidance information ascending.

**P-Sort**: The guidance information of this operator considers both of the priority and requirement image duration. And let a ground target $gt_i$ as an example, the guidance information is defined as (22).

$$GF\_P(gt_i) = \frac{gt_i.d_0}{gt_i.\omega} \qquad (22)$$

where $gt_i.d_0$ and $gt_i.\omega$ represents the requirement image duration and the priority of $gt_i$ respectively.

According to the second knowledge (Knowledge 2) in our previous research (Chang et al., 2020), this operator ranks all selected ground targets by the guidance information ascending, which denotes the ground target with higher priority and shorter requirement image duration will be scheduled in priority.

**E-Sort**: The guidance information of this operator considers the observation energy consumption by requirement image duration and original transition energy consumption. And let a ground target $gt_i$ as an example, the calculation function for generating the guidance information is shown as (23).

$$GF\_E(gt_i) = ROE(gt_i) + OTE(gt_i) \qquad (23)$$

where $ROE(gt_i)$ and $OTE(gt_i)$ represents the observation energy consumption by requirement image duration and original transition energy consumption of $gt_i$ respectively. And they are defined as follows respectively.

$$ROE(gt_i) = gt_i.d_0 \times eo \qquad (24)$$

where $eo$ represents energy consumption rate for observation and $gt_i.d_0$ indicates the requirement image duration of $gt_i$

$$OTE(gt_i) = ec \times trans(\Delta g_{o \to gt_i}) \qquad (25)$$

where $ec$ denotes energy consumption rate for attitude conversion. $o = \{0,0,0\}$ represents the original attitude status of AEOS that the three attitude angles (pitch, roll, and yaw) are all zero. $trans(\Delta g_{o \to gt_i})$ represents the transition time spent by changing attitude angles from the original attitude status to the attitude angles for observing the ground target $gt_i$.

Note that, this operator ranks all selected ground targets by the guidance information ascending, which denotes the ground target consuming less energy will be scheduled in priority.

**C-Sort**: The guidance information of this operator considers the original workpiece congestion defined in our previous research (Chang et al., 2020). Let a ground target $gt_i$ as an example, the original workpiece congestion is defined as the function (26).

$$gt_i.c_0 = \sum_{j=1}^{n} NoD(gt_j.\omega \times d_{ij}) \tag{26}$$

where $d_{ij}$ denotes the conflict distance between $gt_i$ and $gt_j$. If there is an irreconcilable conflict between them, i.e. either-or conflict, $d_{ij} = 1$; if there is a reconcilable conflict, $d_{ij} = 0.5$; else if there is no conflict, $d_{ij} = 0$. Note that, $NoD()$ represents dimensionless processing defined as the function (27).

$$NoD(x_i) = \frac{1}{exp(1 - x_i / \max_{j=1,\cdots,n} x_j)} \tag{27}$$

among them $x_i$ is an independent variable, $\max_{j=1,\cdots,n} x_j$ represents the maximum value in the set $\{x_j | j = 1, \cdots, n\}$ and $n$ indicates the size of the set.

According to the first knowledge (Knowledge 1) in our previous research (Chang et al., 2020), this operator ranks all selected ground targets by the guidance information ascending, which denotes the ground target with less original workpiece congestion will be scheduled in priority.

### 4.2.2. Dichotomization scheduling

| Algorithm 2 : The flow of dichotomization scheduling |
|---|
| Step 1    Let $TB = HB$ for $gt_i$, and justify all constraints. If satisfied, end the scheduling; Otherwise let $TB = EB$ for $gt_i$, and justify all constraints. If satisfied, set $LB = HB$ and update $HB = \frac{LB+EB}{2}$ and go to Step 2; Otherwise let $TB = LB$ for $gt_i$, and justify all constraints. If satisfied, set $EB = HB$ and update $HB = \frac{LB+EB}{2}$ and go to Step 3, otherwise end the scheduling. |
| Step 2    Allocate let $TB = HB$ for $gt_i$, and justify all constraints. If satisfied, set $EB = HB$, otherwise set $LB = HB$. And then update $HB = \frac{LB+EB}{2}$. If $EB = LB$ end the scheduling, otherwise repeat Step 2. |
| Step 3    Allocate let $TB = HB$ for $gt_i$, and justify all constraints. If satisfied, set $LB = HB$, otherwise set $EB = HB$. And then update $HB = \frac{LB+EB}{2}$. If $EB \geq LB$ end the scheduling, otherwise repeat Step 3. |

In addition, to facilitate to present the flow of dichotomization scheduling, as shown in Algorithm 2, some notations are defined as below.

- $TB$    The true observation begin moment of observation window (OW)
- $HB$    The observation begin moment of OW with highest image quality
- $EB$    The earliest observation begin moment of current valid VTW
- $LB$    The latest observation begin moment of current valid VTW

Let a ground target $gt_i$ as an example, at beginning of dichotomization scheduling, the value of these four moments is given as: $TB = Null$, $HB = gt_i.w.b_0 - \frac{gt_i.d_0}{2}$, $EB = gt_i.w.s$ and $LB = gt_i.w.e - gt_i.d_0$.

We will justify the OW with the highest image quality of $gt_i$ in priority. If the constraint is satisfied, dichotomization scheduling will end as shown in Step 1. Otherwise, we will use dichotomization to narrow the valid VTW.

### 4.2.3. Expand operator

It is similar to sort operator, four expand operators, **R-Expand**, **P-Expand**, **E-Expand** and **C-Expand**, are designed to expand the image duration[3].

**R-Expand**: The guidance information of this operator is generated randomly. Then the image duration of all scheduled ground targets is expanded on by one ordering by the guidance information ascending.

**P-Expand**: The guidance information of this operator is calculated by the function (22) under replacing $gt_i.d_0$ as $gt_i.d$. Then expand the image duration of all scheduled targets one by one ordering by the guidance information ascending.

**E-Expand**: The guidance information is calculated by the function (23), and $ROE$ and $OTE$ is calculated based on the true OW. Then expand the image duration of all scheduled ground targets one by one ordering by the guidance information ascending.

**C-Expand**: The guidance information is calculated by the function (26). Note that, the workpiece congestion is calculated only considering all scheduled ground targets. Then expand the image duration of them ordered by the guidance information ascending.

### 4.2.4. The flow of PD/LA+NSGA-II

Combining the PD/LA algorithm and NSGA-II is to generate many solutions and improve schemes. During the combo, the main function of PD/ LA is to generate offspring solutions while that of NSGA-II is to select and save elitist solutions fast.

---

[3] Before expanding, all scheduled ground targets are sorted by their true observation begin moment ascending

## 4.3. Adaptive large neighborhood search metaheuristic algorithm (ALNS)

As mention above, both of PD and LA are greedy constructive algorithms, and they can obtain a feasible or near-optimal solution stably. But they will spend much time, especially embedding them into NSGA-II. On the other hand, the solutions obtained by them are similar to be some extent and the diversity of their population is low. We will discuss these phenomena in the simulation experiments in detail.

Considering both of quality and diversity of solution, we'd like to adopt an adaptive large neighborhood search metaheuristic algorithm (ALNS). The ALNS algorithm can search for good quality solutions, which has been used in several studies (He et al., 2019a, He et al., 2018, Liu et al., 2017) to solve OSPFAS. The basic structure of ALNS is double-loop. The inner loop is a local search process, while the outer loop uses some criteria to control the search process.

### 4.3.1. A random greedy heuristic algorithm (RGHA)

Since ALNS is not highly sensitive to the initial solution and the solution obtained by heuristic greedy is feasible and stable, we propose an initialized algorithm based on the random greedy heuristic (RGHA). The pseudocode of RGHA is shown as Algorithm 3.

| Algorithm 3 : A random greedy heuristic algorithm (RGHA) |
| --- |
| **Input:** A set of ground targets $IGT$, the probability ($BMR$) of each target observed during the OW with the best image quality |
| **Output:** An initial solution $S$, the element of which is a set of ground targets $GT$ which are scheduled to be observed |
| 1: Sort $IGT$ by priority descending and allocate requirement image duration ($d_0$) to all ground targets. |
| 2: **Repeat** ------ **Confirming observed moment** |
| 3:     Choose a ground target $gt$ from $IGT$. |
| 4:     Rand a probability $\tau$, if $\tau$ is bigger than $BMR$, set $gt.w.b_0 - \frac{gt.d_0}{2}$ as the observation begin time of $gt$, otherwise rand observation begin moment belongs to $[gt.w.s, gt.w.e]$. |
| 5:     Estimate whether $gt$ can be observed according to all constraints (18)-(21), if the result is false, abandon $gt$ directly. Otherwise add $gt$ into $GT$. |
| 6: **Until** all ground targets in $IGT$ are visited. |
| 7: **Return** the initial solution $S$. |

The set $IGT$ are is selected from the set $AGT$ adopting the method in section 4.1. In the line 4:, OW with the highest image quality is allocated to every ground target in priority. And we do not consider moving observation moment backward or forward in **dichotomization scheduling** of PD/LA. So, RGHA can obtain many different solutions fast.

### 4.3.2. "Destroy" operators and "Repair" operators

Considering the characteristic of OSWVID, we design two types "Destroy" operators and two types "Repair" operators. "Destroy" operator is used to delete some observed ground targets from a given solution (**Delete operators**) or short image duration of some scheduled ground targets in the given solution (**Short operators**). All deleted ground targets and shorted ground targets are saved into a set called taboo bank $B$ with a given size $|B|$. The set of deleted ground targets is named $DB$ while the set of shorted ground targets is named $SB$, and the size of them is related to their latest weight. We will define the method for calculating and updating the weight in section 4.4. And "Repair" operators are used to inserting some unscheduled ground targets into a given solution (**Insert operators**) or expand the image duration of some scheduled ground targets (**Expand operators**[4]). By the way, some new solutions are generated.

4.3.2.1. Delete operators

Delete operators remove some scheduled ground targets from a given solution. All deleted ground targets are saved into the set $DB$. $DB$ is empty at the beginning of every iteration and filling $DB$ to the full is the termination condition for delete operators. Four different delete operators are defined as follows.

**R-Delete**: This operator deletes some scheduled ground targets from a given solution randomly.

**P-Delete**: The guidance information of this operator is calculated by the function (22) under replacing $gt_i.d_0$ as $gt_i.d$, while **P-Delete** will remove some ground targets with lower priority and longer image duration from the given solution.

**E-Delete**: The guidance information of this operator is the same as that of **E-Sort**, while **E-Delete** will remove some ground targets consuming more energy from the given solution.

**C-Delete**: The guidance information of this operator is the same as that of **C-Sort**, while **C-Delete** will remove some ground targets with larger original workpiece congestion from the given solution.

4.3.2.2. Short operators

Short operators short the image duration of some scheduled ground targets in a given solution. All shorted ground targets are saved into the set $SB$. $SB$ is empty at the beginning of every iteration and filling $SB$ to the full is the termination condition for short operators. Four different short operators are defined as follows.

**R-Short**: This operator shorts the image duration of scheduled ground targets in a given

---
[4] **Expand operators** have been defined in section 0, so we do not repeat them here.

solution randomly.

**P-Short**: The guidance information of this operator is the same as that of **E-Expand**, while **P-Short** will short the image duration of some ground targets with lower priority and longer image duration in the given solution.

**E-Short**: The guidance information of this operator is the same as that of **E-Expand**, while **E-Short** will short the image duration of some ground targets consuming more energy in the given solution.

**C-Short**: The guidance information of this operator is the same as that of **C-Sort**, while **C-Short** will short the image duration of some ground targets with larger original workpiece congestion in the given solution.

4.3.2.3. Insert operators

All unscheduled ground targets are saved in a set denoted by $F$. The ground targets in $F$ but not in $B$ ($DB$ and $SB$) could be selected and inserted into a given scheme to produce a new solution. RGHA is adopted by Insert operators to insert selected ground targets into a given solution. Four different insert operators are defined as follows.

**R-Insert**: This operator selects some unscheduled ground targets in $F$ but not in $B$ randomly and inserts them into the given solution.

**P-Insert**: The guidance information of this operator is the same as that of **P-Sort**, while **P-Insert** will insert some unscheduled ground targets with higher priority and shorter image duration into the given solution.

**E-Insert**: The guidance information of this operator is the same as that of **E-Sort**, while **E-Insert** will insert some unscheduled ground targets consuming less energy into the given solution.

**C-Insert**: The guidance information of this operator is the same as that of **C-Sort**, while **C-Insert** will insert some unscheduled ground target with less original workpiece congestion into the given solution.

### 4.3.3. The flow of ALNS+NSGA-II

Like PD+NSGA-II and LA+NSGA-II, combining the ALNS algorithm and NSGA-II is to generate many solutions and improve the scheme. During the combo, the main function of ALNS is to generate offspring solutions while that of NSGA-II is to select and save elitist solutions fast. Note that, ALNS adopts "Destroy" operators and "Repair" operators based on current elitist solutions to generate offspring solutions.

## 4.4. An adaptive layer and terminational condition

Each operator has a score and a weight. The score depends on the past performance of observation scheduling, and the weight is updated according to the score. Here, the four scores are defined as follows.

- $\sigma_1$ if the new solution dominates all current solutions;
- $\sigma_2$ if the new solution dominates one of the current non-dominated solutions;
- $\sigma_3$ if the new solution on the current Pareto frontier;
- $\sigma_4$ if the new solution is dominated by one of the current non-dominated solutions.

At the end of the iteration, the weights are updated:

$$\omega_i^\alpha = (1-\lambda)\omega_i^\alpha + \lambda \frac{\pi_i^\alpha}{\sum_{i=1}^{|I_\alpha|} \pi_i^\alpha} \qquad (28)$$

where, $\alpha$ denotes the type of operator. $\alpha = \{\text{Delete, Short, Insert, Sort, Expand}\}$, $|I_\alpha|$ represents the number of operators in different types. $\pi_i^\alpha$ and $\omega_i^\alpha$ denotes the score and weight of according to the operator. $\lambda \in [0,1]$ is a reaction factor that controls how sensitive the weights are to changes in the performance of operators, and we will discuss the value of it in section 5.3.

The roulette wheel mechanism is used to choose operator. The utilized rate ($r_i^\alpha$) is calculated according to the following equation.

$$r_i^\alpha = \frac{\omega_i^\alpha}{\sum_{i=1}^{|I_\alpha|} \omega_i^\alpha} \qquad (29)$$

A maximum number of iterations ($MaxIter$) is only one terminational condition to stop the evolution of these three algorithms. $MaxIter$ is given before every evolution.

## 5. Simulation experiments

In this section, the performance of all proposed algorithms will be discussed. Firstly, three simulation experiments are designed to analyze parameter optimization and the evolution of operators. Then, we will compare the efficiency of our three evolutionary algorithms. All algorithms are coded in C#, using Visual Studio 2013, and performed our experiments on a laptop with intel(R) Core (TM) i7-8750H CPU @ 2.2GHz and 16 GB RAM. The general parameters in our three algorithms as shown in Table 1.

Table 1 General parameters

| Parameter | Meaning | Value |
|---|---|---|
| $NS$ | The population size of all solutions | 100 |
| $NBest$ | The population size of elitist solutions | 50 |
| $NA$ | The population size of archive solutions | 100 |
| $MaxIter$ | The maximum number of iterations | 200 |
| $RS$ | The probability for selecting a ground target from $AGT$ | 0.3 |
| $BMR$ | The probability of each ground target observed in best image quality | 0.7 |
| $\sigma_1$ | if the new solution dominates all current solutions; | 30 |
| $\sigma_2$ | if the new solution dominates one of the current non-dominated solutions; | 20 |
| $\sigma_3$ | if the new solution on the current Pareto frontier; | 10 |
| $\sigma_4$ | if the new solution is dominated by one of the current non-dominated solutions. | 0 |
| $\lambda$ | The value of the reaction factor to control update the weight of operators | 0.5 |

The instances, Chinese area distribution(CD) and Worldwide distribution (WD), designed in the research (Liu et al., 2017) are suitable for testing the algorithms for OSPFAS. Since OSWVID is a novel OSPFAS, we adopt their instances to test our three algorithms.

### 5.1. The efficiency analysis for three algorithms

All instances proposed in the research (Liu et al., 2017) are adopted to integrate the efficiency of PD+NSGA-II, LA+NSGA-II, and ALNS+NSGA-II. We consider two important criteria, quality and diversity of obtained solutions, to estimate the efficiency of these three algorithms.

#### 5.1.1. The quality of solutions

On the one hand, in order to estimate the quality of the obtained solutions, we propose three important quotas. The first and second quota is the value of two optimization objectives, LR and EC. Since under the condition that the quality of obtained solutions is similar, the less runtime spends, the better algorithm is, the runtime of algorithms as the last quota, and the unit of it is seconds.

Based on all instances, these three quotas of these three algorithms are shown in Figure 3-Figure 5 respectively. Note that, in Figure 3 and Figure 4, the x-axis and y-axis represent the symbol of instances and the value of according to quota respectively. And a simplified boxplot is proposed to indicate the maximum value, average value and minimum value of quotas, the bar denotes the average value, while the horizontal line above and the horizontal line below represents the maximum value and the minimum value of quotas respectively. While in Figure 5, the x-axis and y-axis represent the symbol of instances and runtime respectively. Note that, the higher bar is, the more runtime spent by the algorithm is.

From Figure 3, we find several interesting phenomena. 1) The average value of LR obtained by ALNS+NSGA-II is slightly worse than that obtained by PD+NSGA-II and LA+NSGA-II for all instances, the average value of LR obtained by PD+NSGA-II and LA+NSGA-II is roughly the same. 2) The minimum value of LR obtained by ALNS+NSGA-II is constantly better than that obtained by PD+NSGA-II and LA+NSGA-II for all instances. 3) The difference between the maximum value of LR and the minimum value of LR obtained by ALNS+NSGA-II is the biggest among them. Given all that, we would like to think that the quality of solutions obtained by these three algorithms is about the same only considering the value of LR. In addition, the diversity of solutions obtained by ALNS+NSGA-II may be much greater for the bigger difference. We will talk about that in detail in section 5.1.2.

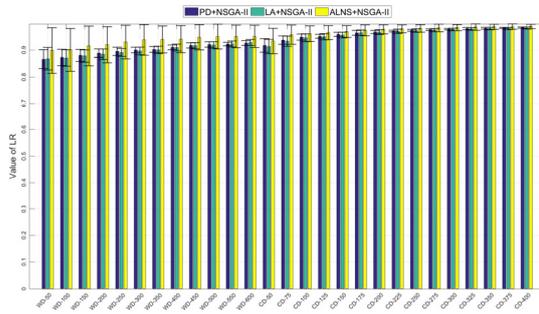

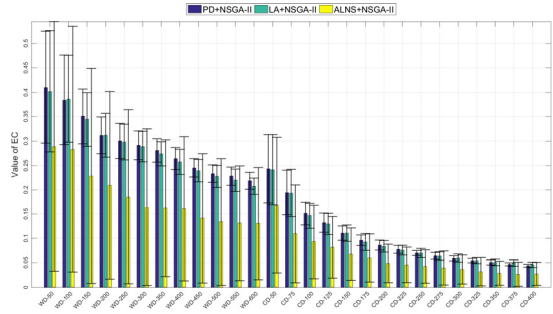

Figure 3 The value of the first optimization objective, LR, for all instances obtained by these three algorithms respectively

Figure 4 The value of the second optimization objective, EC, for all instances obtained by these three algorithms respectively

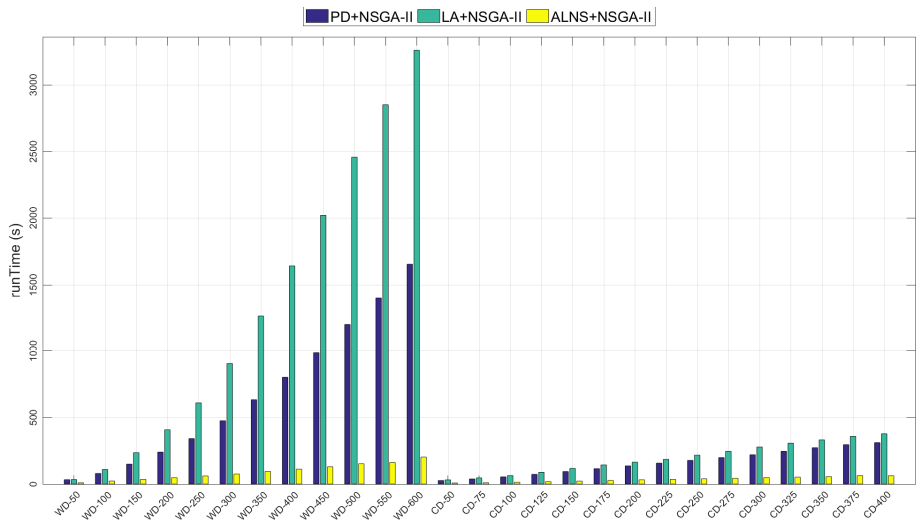

Figure 5 The runtime spent by these three algorithms respectively for all instances obtained

EC of solutions obtained by ALNS+NSGA-II is less than that obtained by PD+NSGA-II and LA+NSGA-II, because the bar of ALNS+NSGA-II is shorter than those of PD+NSGA-II and LA+NSGA-II as shown in Figure 4. The minimum value of EC obtained by ALNS+NSGA-II is also significantly better than that obtained by PD+NSGA-II and LA+NSGA-II for all instances. On the other hand, the maximum value of EC obtained by these three algorithms is roughly the same. So, we would like to think that the quality of solutions obtained by the ALNS+NSGA-II algorithm is best only considering the value of EC. The difference between the maximum value of EC and the minimum value of EC obtained by ALNS+NSGA-II is also the biggest among these three algorithms.

The runtime spent by ALNS+NSGA-II is apparently less than that spent by PD+NSGA-II and LA+NSGA-II shown in Figure 5. Especially with the scale of ground targets increasing, the runtime spent by PD+NSGA-II and LA+NSGA-II raises significantly, while the runtime spent by ALNS+NSGA-II changes stably. So, it is no doubt that the quality of solutions obtained by the ALNS+NSGA-II algorithm is best only considering runtime.

Given all that, the ALNS+NSGA-II algorithm can obtain better solutions within less runtime for all instances. The PD+NSGA-II algorithm and LA+NSGA-II algorithm may only be suitable for the instance with a small scale of ground targets, like WD-50~WD-100 and CD-50~CD-100.

**5.1.2. The diversity of solutions**

The diversity of obtained solutions is important for evolutionary algorithms (Deb et al., 2002). As mentioned above, the diversity of solution obtained by ALNS+NSGA-II is much greater because the difference of the two objectives value obtained by ALNS+NSGA-II is the biggest among these three algorithms. To explain this result further, we choose six instances from the Worldwide distribution (WD), WD-100 ~ WD~600 steps 100, as examples. The Pareto frontier of them obtained by these three algorithms is shown in Figure 6, and subplot 1 - subplot 6 represents the Pareto frontier of them respectively.

(1) The Pareto frontier obtained by PD+NSGA-II and LA+NSGA-II is most located above that obtained by ALNS+NSGA-II according to subplot 1-subplot 6. It means the quality of solutions obtained by PD+NSGA-II and LA+NSGA-II is worse than that obtained by ALNS+NSGA-II, which is consistent with the result mentioned in section 5.1.1.

(2) The diversity of solutions obtained by ALNS+NSGA-II is much bigger than that of others because the length of the Pareto frontier obtained by ALNS+NSGA-II is significantly longer than others.

(3) The much longer Pareto frontier results in the larger average value of optimization objectives, as mentioned in section 5.1.1, so the average value cannot reflect the quality of solutions directly.

(4) Considering the quality and the diversity of obtained solutions together, the ALNS+NSGA-II algorithm is constantly better than the PD+NSGA-II algorithm and the LA+NSGA-II algorithm for all instances. The PD+NSGA-II algorithm and the LA+NSGA-II algorithm are roughly same, but the LA+NSGA-II algorithm is much slower especially for the large-scale instances.

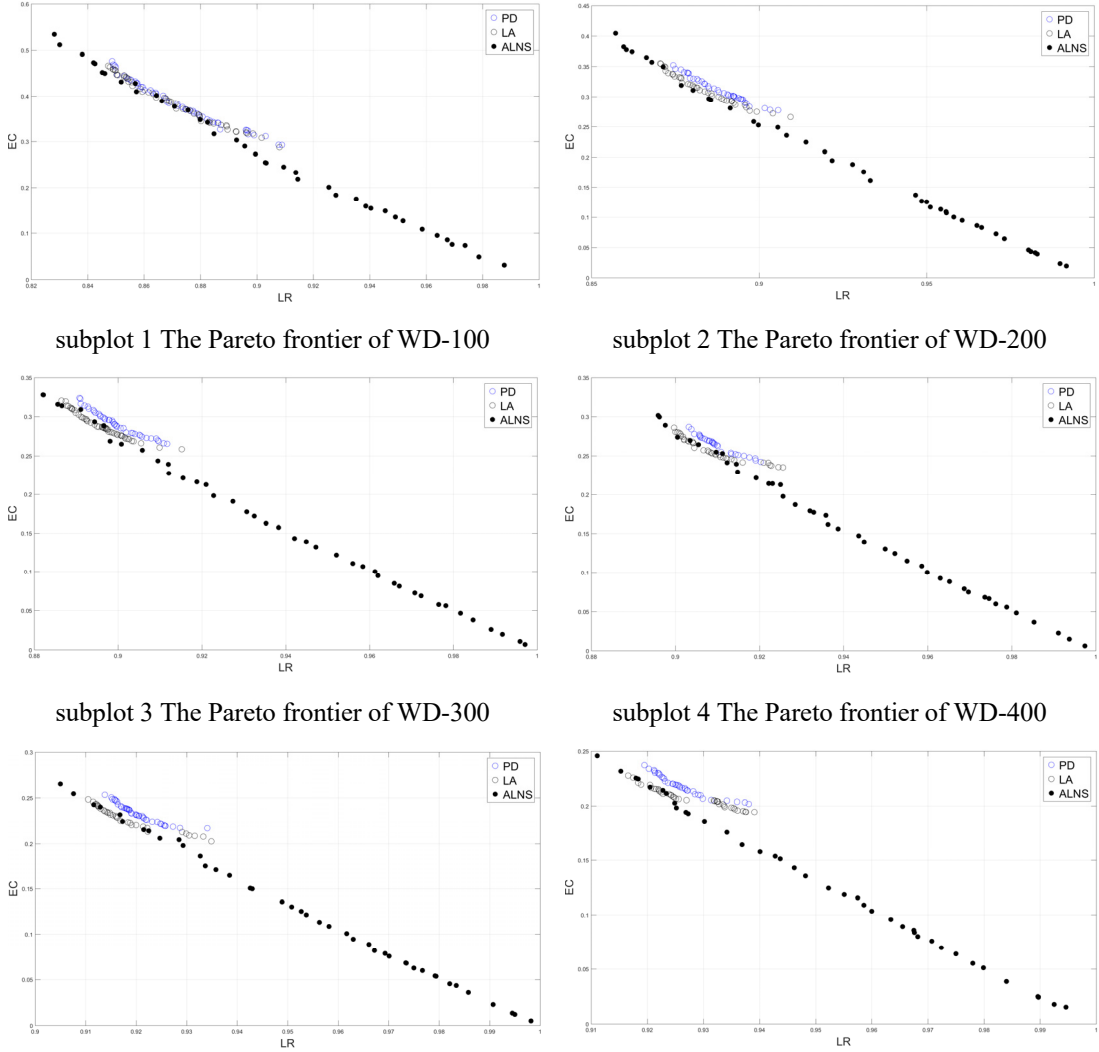

subplot 1 The Pareto frontier of WD-100

subplot 2 The Pareto frontier of WD-200

subplot 3 The Pareto frontier of WD-300

subplot 4 The Pareto frontier of WD-400

subplot 5 The Pareto frontier of WD-500

subplot 6 The Pareto frontier of WD-600

Figure 6 The Pareto frontier of WD-100 ~ WD~600 obtained by PD+NSGA-II, LA+NSGA-II and ALNS+NSGA-II respectively

## 5.2. Parameter optimization of $RS$

CD-50, 50 ground targets in Chinese area distribution, is chosen to optimize the parameter $RS$. Let $RS$ belongs to the interval $[0, 0.9]$[5] and steps 0.1. The elitist solutions after 200 iterations obtained by these three algorithms under different values of $RS$ are shown in Table 2. $\widehat{v_1}$, $\overline{v_1}$ and $\widetilde{v_1}$ represents maximum, average and minimum of LR, while $\widehat{v_2}$, $\overline{v_2}$ and $\widetilde{v_2}$ denotes maximum, average, and minimum of EC.

---

[5] If $RS = 1$, any ground target can selected from the set $AGT$, so we do not consider that value.

Table 2 The elitist solutions obtained by the three algorithms under different value of RS

| RS | PD+NSGA-II | | | | | | | | LA+NSGA-II | | | | | | | | ALNS+NSGA-II | | | | | | | |
|---|---|---|---|---|---|---|---|---|---|---|---|---|---|---|---|---|---|---|---|---|---|---|---|---|
| | $\widetilde{v}_1$ | $\overline{v}_1$ | $\widehat{v}_1$ | $\widetilde{v}_2$ | $\overline{v}_2$ | $\widehat{v}_2$ | $\widetilde{v}_1$ | $\overline{v}_1$ | $\widehat{v}_1$ | $\widetilde{v}_2$ | $\overline{v}_2$ | $\widehat{v}_2$ | $\widetilde{v}_1$ | $\overline{v}_1$ | $\widehat{v}_1$ | $\widetilde{v}_2$ | $\overline{v}_2$ | $\widehat{v}_2$ |
| 0   | 0.9009 | 0.9097 | 0.9299 | 0.2395 | 0.2776 | 0.3178 | 0.8972 | 0.9085 | 0.9309 | 0.2331 | 0.2557 | 0.3161 | 0.8976 | 0.9387 | 0.9879 | 0.0414 | 0.1506 | 0.2777 |
| 0.1 | **0.8997** | **0.912** | **0.9347** | **0.1913** | **0.2539** | **0.3043** | **0.8976** | **0.9077** | **0.9413** | **0.1956** | **0.2506** | **0.2983** | **0.8904** | **0.9361** | **0.9932** | **0.0261** | **0.16** | **0.3021** |
| 0.2 | 0.9009 | 0.918  | 0.9488 | 0.1634 | 0.2364 | 0.3153 | 0.8976 | 0.9145 | 0.9536 | 0.161  | 0.2324 | 0.296  | 0.8961 | 0.9429 | 0.9925 | 0.0241 | 0.1397 | 0.2877 |
| 0.3 | 0.9013 | 0.9235 | 0.9589 | 0.1437 | 0.2169 | 0.3145 | 0.8988 | 0.9184 | 0.9557 | 0.1486 | 0.2175 | 0.3185 | 0.9026 | 0.9391 | 0.9926 | 0.028  | 0.156  | 0.2643 |
| 0.4 | 0.9042 | 0.9265 | 0.9579 | 0.1307 | 0.2017 | 0.2847 | 0.9003 | 0.9205 | 0.9591 | 0.1204 | 0.2106 | 0.2983 | 0.9065 | 0.9444 | 0.9924 | 0.0248 | 0.1374 | 0.2564 |
| 0.5 | 0.9023 | 0.9321 | 0.9704 | 0.0907 | 0.1856 | 0.2983 | 0.9033 | 0.9332 | 0.9732 | 0.09   | 0.1761 | 0.274  | 0.9078 | 0.9529 | 0.9992 | 0.0021 | 0.1214 | 0.2503 |
| 0.6 | 0.9075 | 0.943  | 0.9776 | 0.0675 | 0.1575 | 0.282  | 0.906  | 0.9399 | 0.9785 | 0.0684 | 0.1587 | 0.2724 | 0.9167 | 0.9554 | 0.9979 | 0.0067 | 0.1149 | 0.2152 |
| 0.7 | 0.9131 | 0.9537 | 0.9891 | 0.0329 | 0.125  | 0.2562 | 0.9106 | 0.9471 | 0.986  | 0.0391 | 0.1402 | 0.2653 | 0.9249 | 0.9581 | 0.9991 | 0.0021 | 0.1074 | 0.2065 |
| 0.8 | 0.9235 | 0.9661 | 0.9983 | 0.0045 | 0.0865 | 0.2286 | 0.9223 | 0.9649 | 0.9983 | 0.0038 | 0.0892 | 0.2226 | 0.927  | 0.9575 | 0.9987 | 0.0028 | 0.1164 | 0.2288 |
| 0.9 | 0.9403 | 0.9724 | 0.9995 | 0.0027 | 0.0688 | 0.1672 | 0.9359 | 0.9705 | 0.9995 | 0.0027 | 0.073  | 0.1858 | 0.9552 | 0.9755 | 0.9991 | 0.0028 | 0.0637 | 0.1365 |

According to Table 2, when $RS$ equals {0, 0.1, 0.2, 0.3}, the value of LR and EC achieves a much better balance. The loss rate of image quality (LR) is relatively less, while the energy consumption (EC) is also not very large. It means the quality of solutions obtained by the three algorithms under the value of $RS$ equals {0, 0.1, 0.2 and 0.3} is much better.

On the other hand, the diversity of solutions is another important indicator to estimate the multi-optimization algorithm. The no-dominated solutions obtained by these three algorithms under $RS$ equals these four different values, {0, 0.1, 0.2, 0.3}, are shown in Figure 7-Figure 9.

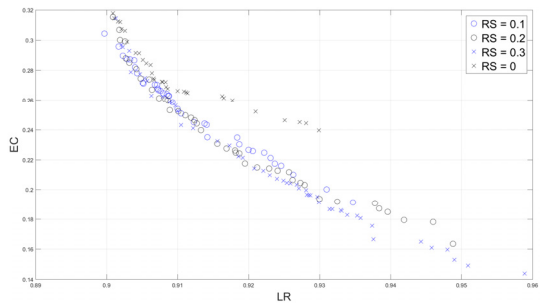
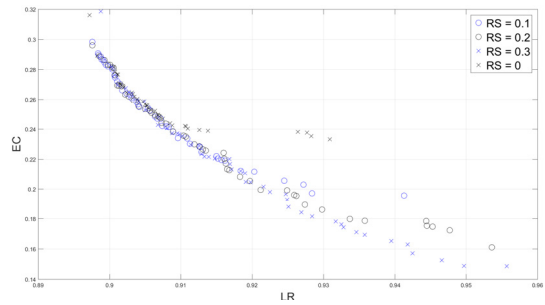

Figure 7 The Pareto frontiers obtained by PD+NSGA-II under four different values of RS

Figure 8 The Pareto frontiers obtained by LA+NSGA-II under four different values of RS

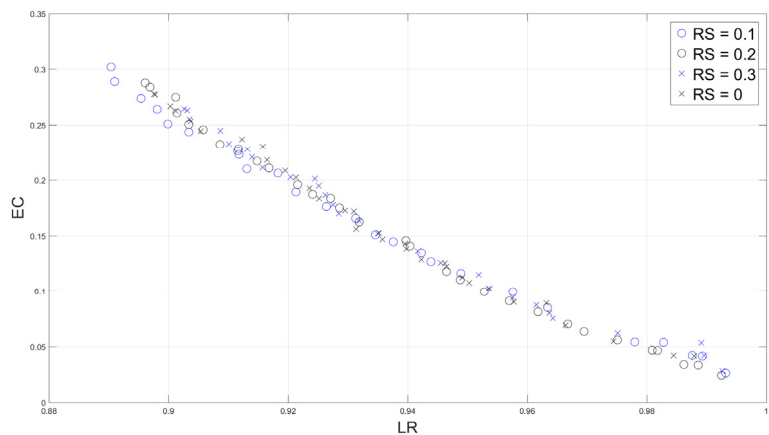

Figure 9 The Pareto frontiers obtained by ALNS+NSGA-II under four different values of RS

Toward PD+NSGA-II and LA+NSGA-II, as shown in Figure 7 and Figure 8, it is apparent that when $RS$ equals zero[6], the obtained Pareto frontier is worst. Because most non-dominated solutions under $RS = 0$ are located in the dominated region under other three values. The Pareto frontier obtained under $RS = 0.1$, $RS = 0.2$ and $RS = 0.3$ are similar, while that under $RS = 0.2$ is relatively longest among them. So, considering both of quality and diversity of obtained solutions, we suggest setting $RS = 0.2$ for adopting PD+NSGA-II and LA+NSGA-II.

The phenomenon of ALNS+NSGA-II is simple, as shown in Figure 9, both the quality and diversity of obtained solutions achieve best when $RS = 0.1$. So, it is no doubt that $RS = 0.1$ is the best choice for adopting ALNS +NSGA-II.

---

[6] All ground targets in $AGT$ are selected for scheduling

## 5.3. Parameter optimization of $\lambda$

WD-100, 100 ground targets in Worldwide distribution, is chosen to optimize the parameter $\lambda$. Let $\lambda$ belongs to the interval [0,1] and steps 0.1. After 200 iterations, the final weight of all operators within the according algorithm under different $\lambda$ is shown in Figure 10-Figure 12.

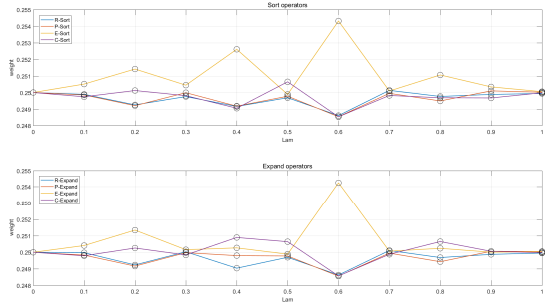
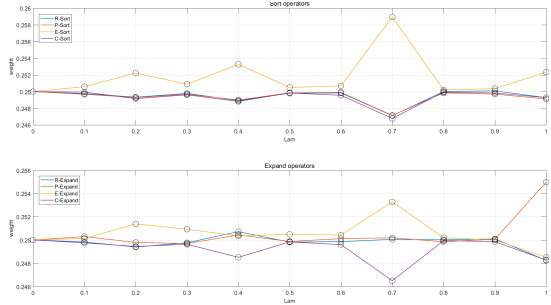

Figure 10 The final weight of Sort/Expand operators used in PD+NSGA-II under different value of $\lambda$

Figure 11 The final weight of Sort/Expand operators used in LA+NSGA-II under different value of $\lambda$

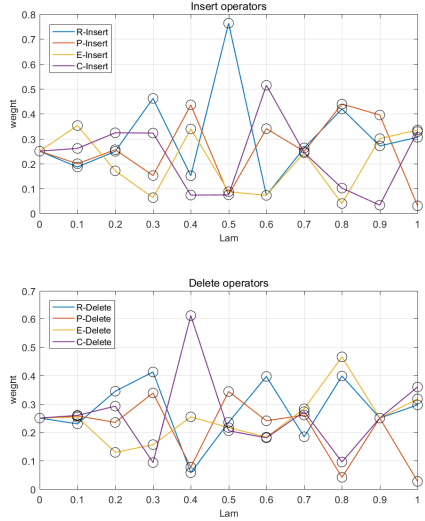
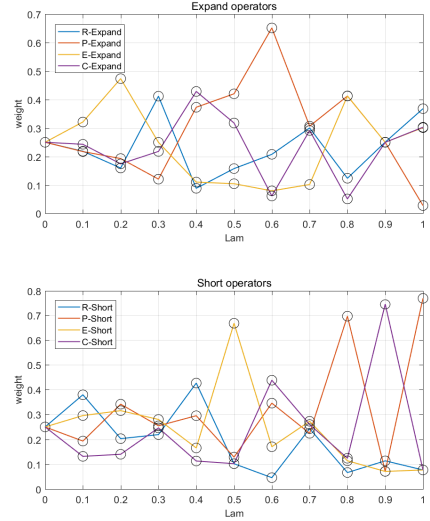

Figure 12 The final weight of Insert/Expand/Delete/Short operators used in ALNS+NSGA-II under different value of $\lambda$

For the PD+NSGA-II algorithm, the final weight of all operators, except **E-Sort/E-Expand**, changes stably according to different $\lambda$ as shown in Figure 10. When $\lambda$ equals 0.6, the final weight of **E-Sort/E-Expand** achieves maximum. But in order to make sure more operators can contribute to the PD+NSGA-II algorithm, which will be good for keeping the diversity of solutions, we suggest setting $\lambda = 0.7$ for adopting PD+NSGA-II in the following experiments.

There is a similar phenomenon toward the LA+NSGA-II algorithm, the final weight of **E-Sort/E-Expand** changes more dramatically according to different $\lambda$ as shown in Figure 11. The final weight of **E-Sort** is constantly better than that of others. When $\lambda$ equals 0.7, the final weight of **E-Sort/E-Expand** achieves maximum. While the final weight of expanding operators changes disorderly. But when $\lambda$ equals 0.5, the final weight of all operators is nearest. So, we will set $\lambda = 0.5$ for adopting LA+NSGA-II in the following simulation experiments.

For the ALNS+NSGA-II algorithm, the final weight of all Insert/Expand/Delete/Short operators changes disorderly under different $\lambda$ as shown in Figure 12.

(1) Toward the four Insert operators, when $\lambda$ is less than 0.5, the final weight of **R-Insert** is constantly better. When $\lambda$ belongs to [0.6,1], the final weight of all Insert operators is confused. While $\lambda$ equals 0.7, the final weight of all Insert operators is nearest.

(2) Toward the four Expand operators, when $\lambda$ is less than 0.3, the final weight of **E-Expand** is constantly better. When $\lambda$ belongs to [0.4,0.7], the final weight of **P-Expand** is constantly better. But when $\lambda$ is bigger than 0.7, the final weight of all Expand operators is disorderly. While we can find that $\lambda$ belongs to {0.2, 0.7, 0.9}, the final weight of all Expand operators is near.

(3) The final weight of Delete/Short operators is always disorderly when $\lambda$ belongs to the interval [0,1]. But when $\lambda$ equals {0.1, 0.7}, the final weight of all Delete operators is near, while when $\lambda$ belongs to {0.3, 0.7}, the final weight of all Short operators is near.

Given all that, in order to make sure more operators can contribute to the ALNS+NSGA-II algorithm, we suggest setting $\lambda = 0.7$ for adopting ALNS+NSGA-II algorithm in the following simulation experiments.

### 5.4. Operators evolution

WD-50, 50 ground targets in Worldwide distribution, is chosen to analyze the evolution of all operators within the three algorithms. Set $MaxIter$ equals 1000 and $RS$ and $\lambda$ according to the suggestions in section 0 and section 5.3. After 1000 iterations, the mean weight of every operator within these three algorithms is shown in Figure 13-Figure 15 respectively.

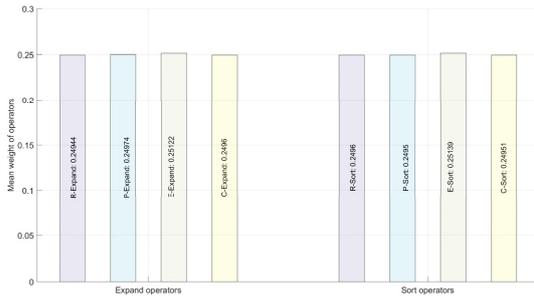

Figure 13 The mean weight of every operator within the PD+NSGA-II algorithm

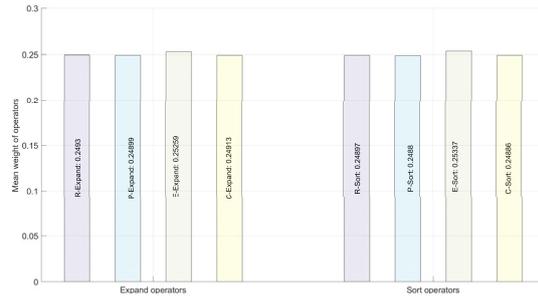

Figure 14 The mean weight of every operator within the LA+NSGA-II algorithm

Toward PD+NSGA-II and LA+NSGA-II, it is consistent with the suggestion as mentioned in section 5.3 that the mean weight of every operator within them is near and around 0.25.

There is a gap between the mean weight of every operator within ALNS+NSGA-II. Like the four operators in Insert operator, the mean weight of **R-Insert** (0.23835) is least and that of **P-Insert** (0.2598) is greatest among them, but the gap of them is very small and they are still around 0.25, which is also consistent with the suggestion as mentioned in section 5.3.

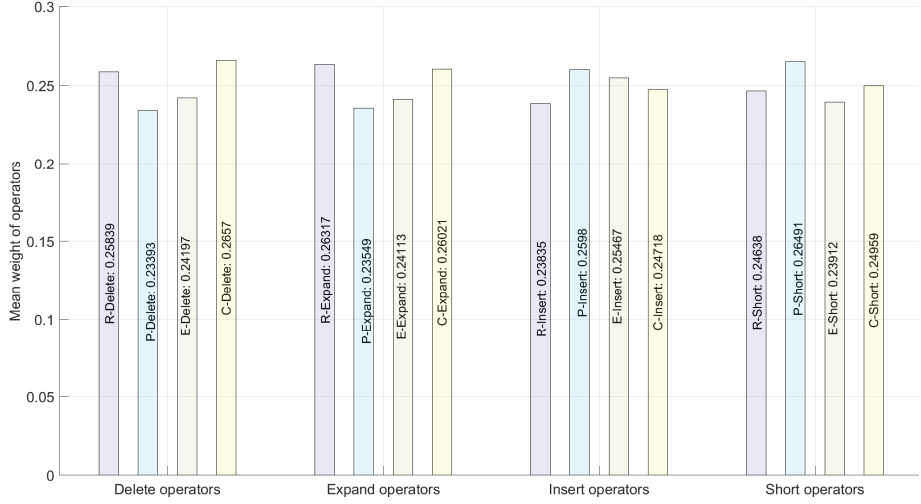

Figure 15 The mean weight of every operator within the ALNS+NSGA-II algorithm

## 6. Conclusion

Active imaging enriches the extension of observation scheduling problem for agile earth observation satellites (OSPFAS). AEOSs with the capability of active imaging can modify their attitude angles during observation, therefore the image duration of each ground target is variable. Since the variable image duration of each ground target, we present a bi-objective optimization model to minimize the loss rate of image quality (LR) and energy consumption (EC) for describing Observation scheduling problem for AEOS with variable image duration (OSWVID). Note that, the image quality is a cumulative variable considering image duration and image attitude angles during the whole observation window (OW). According to previous researches (Deb et al., 2002, Liu et al., 2017, Wolfe and Sorensen, 2000), we propose three multi-objective memetic algorithms, PD+NSGA-II, LA+NSGA-II, and ALNS+NSGA-II, to solve OSWVID. Based on Knowledge 1 and Knowledge 2 summarized in our previous research (Chang et al., 2020), we design several operators, **Sort** operators, **Insert** operators, **Expand** operators, **Delete** operators, and **Short** operators, to improve these three algorithms. In addition, all operators evolve controlled by an adaptive layer.

Based on existing instances designed in the research (Liu et al., 2017), extensive simulation experiments demonstrate the efficiency of PD+NSGA-II, LA+NSGA-II, and ALNS+NSGA-II respectively.

(1) Since the ground targets are redundant (Wang et al., 2016) according to the image capability of AEOS, selecting the ground target to schedule will reflect observation scheduling directly. So, In the first simulation experiment, we optimize the parameter, *RS*, which is used to control the selecting process. Considering the quality and diversity of obtained solutions, we suggest setting $RS = 0.2$ for adopting PD+NSGA-II and LA+NSGA-II, while setting $RS = 0.1$ for adopting ALNS+NSGA-II.

(2) The value of $\lambda$ is a critical fact that affects the adaptive layer. According to parameter optimization of $\lambda$, setting $\lambda = 0.7$ for adopting PD+NSGA-II and ALNS+NSGA-II, and setting $\lambda = 0.5$ for adopting LA+NSGA-II will help more operators work in corresponding algorithms. The results of the evolution of operators are consistent with the suggestion obtained in the parameter optimization of $\lambda$.

(3) Based on the whole instances proposed in the research (Liu et al., 2017), we consider the quality and diversity of obtained solutions to estimate the efficiency of these three algorithms. The ALNS+NSGA-II algorithm can obtain better solutions within less runtime for all instances. While the PD+NSGA-II algorithm and LA+NSGA-II algorithm are roughly the same and may only be suitable for the instances with small-scale ground targets.

## Acknowledgements

The research of Zhongxiang Chang was supported by the science and technology innovation Program of Hunan Province (2021RC2048).